# Automatic Generation of Machine Learning Synthetic Data using ROS[*]

Kyle M. Hart[1][0000-0001-8254-5722], Ari B. Goodman[1], and Ryan P. O'Shea[1]

[1] Naval Air Warfare Center – Aircraft Division – Lakehurst, Lakehurst, NJ 08757 USA
kyle.m.hart@navy.mil, ari.b.goodman@navy.mil,
ryan.oshea3@navy.mil

**Abstract.** Data labeling is a time intensive process. As such, many data scientists use various tools to aid in the data generation and labeling process. While these tools help automate labeling, many still require user interaction throughout the process. Additionally, most target only a few network frameworks. Any researchers exploring multiple frameworks must find additional tools or write conversion scripts. This paper presents an automated tool for generating synthetic data in arbitrary network formats. It uses Robot Operating System (ROS) and Gazebo, which are common tools in the robotics community. Through ROS paradigms, it allows extensive user customization of the simulation environment and data generation process. Additionally, a plugin-like framework allows the development of arbitrary data format writers without the need to change the main body of code. Using this tool, the authors were able to generate an arbitrarily large image dataset for three unique training formats using approximately 15 minutes of user setup time and a variable amount of hands-off run time, depending on the dataset size. The source code for this data generation tool is available at https://github.com/Navy-RISE-Lab/nn_data_collection

**Keywords:** Machine Learning, Data Generation, ROS

## 1 Introduction

Data labeling is such a time intensive and tedious task that many data scientists turn to existing datasets or outsource their data labeling process to others. However, there are occasions where scientists may require brand new datasets due to mission needs. Creating these custom datasets is a time-consuming process due to the time involved with data acquisition and data labeling. The time further increases if the researchers explore different networks or new scenarios after collecting the dataset. For example, if researchers wish to investigate scenes with different objects, they must create and label more data containing these new objects. Furthermore, researchers must make

---

sure to account for any potential statistical changes in their dataset when they add or remove data. Often due to limited time and financial resources, researchers must be careful to select parameters to succinctly create their dataset. As the number of different parameters increase, so too does the dataset size and therefore the cost as well.

Because of this, a variety of tools are available to aid with the data labeling process. These tools assist in the labeling process with varying levels of automation. This work introduces a new fully automated tool targeted towards the creation of data sets using Robot Operating System (ROS) [1] and Gazebo [2], which are common tools within the robotics community. Using ROS and Gazebo, this tool generates perfectly labeled data from arbitrary user-specified scenes. By using common robotic toolsets, this tool provides a familiar means by which robotics researchers can build labeled image datasets for their work.

This tool is designed for minimal user involvement during the data generation process and maximum flexibility. The user can specify arbitrary motion plans for various objects in the scene, the camera position, simulated frame rate, and more. Additionally, the tool allows the addition of arbitrary new data formats through a plugin-like structure. This provides extensibility to new use cases not originally considered. Once configured, the tool automatically runs without requiring user involvement, dramatically reducing hands-on time.

The paper structure is as follows. First is an overview of other data labeling tools. Then, the authors describe their first version of this tool, which focused on a specific use case. Next, the paper describes the improved version of the tool. A discussion on usability follows, including steps required by the user, information on creating new data format writers, and a small vignette to illustrate the tool's effectiveness. Lastly, the authors identify several follow-on steps to further improve the tool.

Additionally, the source code for the tool described here is available at https://github.com/Navy-RISE-Lab/nn_data_collection

## 2    Related Work

There is a wide array of techniques used to obtain labeled data for neural network training. These include the use of existing labeled datasets, leveraging data augmentation techniques on unlabeled data, and manually creating and labeling new datasets. Roh et. al. [3] provide a comprehensive overview of the various techniques. Many current techniques use existing datasets for training. However, there are cases where a project requires the creation of a new dataset through manual image labeling.

The traditional way to create labeled datasets involves manually labeling each image. There are several tools, such as Yolo_mark [4] and LabelImg [5], that assist a user with labeling. These tools typically provide user interfaces to aid in quickly drawing bounding boxes and assigning labels to an image. The tools then write the labels to file in a specified format. While these tools speed up the data labeling process, they still require the user to draw the boxes and assign the labels or require user confirmation that an automated guess is correct. Depending on the size of the dataset, labeling can be a laborious process. For example, the authors used Yolo_mark to label a video approximately a minute and a half in length in about three hours. The process can also introduce human error into the dataset if an image is accidentally mislabeled. Ideally, multiple people will label each image; however, having multiple people label each image significantly increases the amount of time and money involved in creating the dataset. Additionally, these tools typically target only one or a few networks. Extending datasets to additional networks in the worst case can be impossible without relabeling missing data, and in the best case requires deep knowledge of the tool's code or a post-processing conversion script.

There is also a growing body of research on using neural networks to create and label data for training and testing other neural networks. Sixt, et al. [6] and Pfeiffer et al. [7] both use generative adversarial networks (GANs) to take ideal 3D models of objects of interest and place them into realistic synthetic images. They use example images to mimic the lighting conditions, blur, and other characteristics. Similarly, Lee at al. [8] use a neural network to perform style transfer. These approaches allow for automatically generating large datasets. However, the examples the authors found were limited to single objects found within an image and focused on image labels instead of multiple objects within an image.

Besginow et al. [9] propose an alternative approach that generates labeled data for a single object. They use a hardware setup to capture an object from multiple angles. They also supply simple interfaces to allow some user customization for a semi-automated approach. The tool then generates the object detection labels. However, this approach is limited to objects that fit within the hardware setup. Additionally, they only target a single network output format.

Additionally, Dutta et al. [10] provide an overview of a number of automatic labeling methods that leverage machine learning techniques. This includes nearest neighbor approaches, neural network approaches, and SVM classifiers. The nearest neighbor approaches use already labeled images to match based on visual similarity. This requires a sufficiently diverse collection of labeled images, which may not exist for new datasets. All approaches also potentially result in some mislabeling, depending on the accuracy of the model used.

Some researchers have also begun to use other sensor modalities to automatically label training data. Kuhner et al. [11] propose the use of LiDAR in driverless cars to automatically create semantic labels for the image data generated by the car's cameras. Their approach quickly annotates images of roads and curbs for use in training neural networks used to detect and navigate around these objects. This approach offers fast generation of labeled real world data but does not generalize to other problem domains.

Lastly, the authors were unable to find any examples of data generation tools using ROS and Gazebo for scene simulation. Some synthetic generation methods use general purpose simulation environments. For example, Lee et al. [8] use the 3D game engine Unity to generate synthetic training data of wrenches in an industrial environment. Others, like the LGSVL simulator [12], are domain specific.

## 3 Version 1

The first version of this utility originated in a previous project that used YOLOv3 [13] and the Darknet [14] framework. Part of that project explored the impact of several setup parameters, such as image resolution, on network accuracy. This required multiple datasets composed of several simulated robots in arbitrary poses within a simulated environment. Performing this labeling by hand was too time consuming. To speed this up, the team created a ROS package that automatically generated the necessary data in the correct format. By using perfect knowledge of the simulated world, the labels are guaranteed to be accurate. Additionally, the automatic nature of the tool means the authors could run the tool during off hours without the extensive time commitment needed to hand label the data.

The general algorithm is below. The project focused on object detection within single images, so the algorithm uses the entire available space within the simulated environment for robot placement. Additionally, because of the authors' familiarity, they chose to write this version of the tool in C++.

```
Version 1 Procedure:
Load user parameters
For each required datapoint:
  image <- capture new image
  Write to file(image)
  For each object:
    selected_pixel <- select pixel from map pixels    (1)
    If selected_pixel is not free:
      GOTO (1)
    radius <- load user specified safety radius
    outer_pixels <- Bresenham's Circle Algo-
        rithm(selected_pixel, radius)
    For each outer_pixel in outer_pixels:
      line_pixels <- Bresenham's Line Algo-
          rithm(selected_pixel, outer_pixel)
      For each line_pixel in line_pixels:
        If line_pixel is not free:
        GOTO (1)
    position <- transform selected_pixel to coordinate
    orientation <- select value from (-pi, pi]
    Move object in simulation(position, orientation)
    bounding_rectangle <- Project Shape into Image
    Write to label file(bounding_rectangle, class id)
```

To execute the program, the user first specifies several settings via ROS's parameter server, which is a standard way to customize routines in ROS. Specifically, this includes information about where to save files, which robots to use, and each robot's footprint, height, and a safety radius that ensures no accidental collisions during object placement. Additionally, the user must publish a map of the environment indicat-ing which areas are free or occupied. This map also follows a standard ROS convention and uses pixel values to indicate if a space's occupancy status. After providing this information and starting the simulation environment, the user can then run the generation utility and let it auto-generate until finished.

While running, the tool selects a random pose for the robot. To do this, it picks a free pixel on the map, then uses the robot's safety radius, Bresenham's Line drawing algorithm [15], and Bresenham's Circle drawing algorithm [16] to identify if the robot can fit at the selected spot. The drawing algorithms identify which pixels the robot could potentially occupy for a given pose. It then checks if each pixel is already occupied or not. If any pixel is occupied, the algorithm selects a new position, and the process begins again. If the space is free, the algorithm selects a random orientation for the robot. The algorithm then requests Gazebo to move the robot to that pose. This process then repeats for each robot.

After each robot is in position, the tool captures the image data from a simulated camera within the environment, using ROS's publisher/subscriber model. When capturing the image, it also captures the camera information, such as its intrinsic parameters.

Using this information, the tool then generates the labels using the algorithm shown below. It uses the robot's pose and its user-specified shape to construct a rectangular cuboid that circumscribes the robot in the robot's frame of reference. It then transforms the vertices of this cuboid into the image using known positional data and the camera's parameters. Next, it circumscribes a bounding box around the projected vertices using OpenCV [17]. Because these are vertices of a bounding shape, this bounding box is guaranteed to encompass the robot on the image. As proof, consider the case where the object has some feature outside of the bounding rectangle on the image. If this is true, then the vertices projected on the image would also extend outside the rectangle, as the vertices circumscribe the object in 3D space. However, the bounding rectangle is constructed to circumscribe all the projected vertices, which contradicts the original premise of this case. Therefore, this bounding box on the image contains the entire robot within it.

```
Project Shape into Image(object):
vertices <- load user specified object shape
transform <- lookup transform to camera frame(object)
projected_vertices <- apply transform(vertices, trans-
    form)
camera_matrix <- lookup camera parameters
projected_points <- apply camera projec-
tion(projected_vertices, camera_matrix)
bounding_rectangle <-
    OpenCV.BoundingRect(projected_points)
Return bounding_rectangle
```

Lastly, the tool writes the values to file in the correct format. Darknet datasets consist of two elements. The first is the set of raw, unlabeled images. The second element is an associated text file for each image. Each line of the text file contains the infor-mation for one object within the image. This information includes the class id for the class of the object, and the size and location of the bounding box, expressed as fractions of the overall width and height of the image. For example, a bounding box highlighting an object with class id 1, centered on the image with width and height one quarter of the respective width and height for the image would be labeled as *"1 0.5 0.5 0.25 0.25"*

This entire process can then repeat until the tool creates the desired amount of data. Additionally, the user can easily rerun the process with new settings to create validation or test datasets, or to explore new scenarios. Rerunning takes no added user input other than the time required to change settings.

Using this method, the authors were able to generate a fully labeled 1,000 image synthetic dataset with about 10 minutes of setup time and 30 minutes of data generation time. The setup time is the only time that requires user operation and does not increase as the size of the generated dataset increases. The data generation time runs without the user and scales approximately linearly with the number of samples. Once generated, the data was immediately used for network training without any additional post-processing. This allowed the team to quickly explore several elements to the project without spending work time on data creation.

However, there are some drawbacks to this tool. As mentioned previously, the algorithm targets the Darknet format and design considerations, such as moving the robot to arbitrary positions on the map, were made because of it. Any new format would require extensive rewrites. Additionally, there are several required dependencies for the package to work. While some are standard ROS packages found in any ROS installation, some dependencies are less common. Some were even lab specific ROS packages, preventing widespread use of this tool. When the main project began work on a second phase, it became clear that the team required a tool with greater flexibility.

## 4  Version 2

After completion of the first phase of the project, the underlying research expanded to include improved neural network methods. This included new network types, more detailed detection, and the use of video data instead of isolated images. Adapting the existing synthetic generation tool would require significant work. Therefore, the team took the opportunity to redesign the tool to promote greater modularity and reduce the need for future rewrites. The four primary goals for this version were as follows.

- Allow the user to specify and label video frames and sequential images to capture motion between images.
- Use a modular structure to allow the easy creation of new data formats as needed.
- Generate data to train instance segmentation networks.
- Reduce the required dependencies to run the tool, ideally to only a few that are likely to already exist on a computer with ROS and neural network frameworks installed.

Additionally, the team decided to rewrite the entire package in Python. Python is a primary language amongst both data scientists and ROS users, so any potential users are more likely to be familiar with Python than they would be with C++.

### 4.1 Motion Generation

The first version of the generation tool created sets of unrelated images, so the tool moved objects to arbitrary locations at each successive image it generated. The later stage of the project relies on video data fed in as successive frames to exploit state information between images. Therefore, the tool needs to generate data the shows an object's motion across a series of images. To support modularity, this requires the ability to specify any possible motion path in an intuitive way, to allow the exploration of a variety of motion scenarios.

Achieving this flexibility uses a built-in capability of ROS. ROS uses packages called tf and tf2 [18] to track relative pose information between every frame of reference in the scene throughout the entire simulation. This structure is called the TF tree. Additionally, ROS supplies a way to record any information published during simulation using a concept known as bag files. Bag files are a common means by which researchers record their experiments for use later. The algorithm simply reads and stores a TF tree from a pre-recorded bag file to generate a record of each objects pose throughout the simulation.

To generate this bag file, the user can use whatever methods they want. They can use real or simulated robots, joystick control or autonomous control, or any other means. As TF trees are extremely common in ROS robotic simulations, most setups already produce the required information, so it is a simple matter to record the data. Additionally, trees are represented as text information, so the file size is small, even for large simulations.

When run, the algorithm reads in the TF tree from the bag file and uses it as a set of instructions to recreate the scene. It follows the algorithm shown below. Using a user specified frame rate, the algorithm steps through the scene and finds the pose of each object at that moment in time, based on the TF tree. It then moves the objects to those same poses in the Gazebo simulation, recreating the recorded scene. It is important to note that this recreation does not have to occur in real time. Since the bag file contains the positions for the entire simulation, the algorithm can spend as much time processing information as needed before moving on to the next frame, similar to the use of Claymation in movies.

```
Scene Generation Procedure:
Load bag file
start_time <- identify valid start time from bag file
end_time <- identify valid end time from bag file
frame_rate <- load user specified frame rate
current_time <- start_time
While current_time <- end_time:
  For each object to control:
```

```
   pose <- lookup object pose in bag file
   move object in simulation(pose)
scene_data <- Capture all scene data
For each format writer:
   format_writer.WriteScene(scene_data)
current_time <- current_time + (1.0 / frame_rate)
```

An added user benefit of this approach is the ability to create new scenes. When recreating a frame using the TF tree, the algorithm does not care about the visual appearance of the object nor the position of the simulated camera. This means a user can create a single motion plan, then generate data for several different setups. These alternate setups can include different environments, lighting, and camera positions. Alternate setups can also include new types of objects following the same motion paths. The algorithm is agnostic to each and does not require an updated bag file. For instance, the authors developed some example data in an empty environment, then used the same motion plan in an environment with walls and a new camera position. The authors did not need to change the recorded bag file, so created new datasets without any significant user involvement.

This motion recording framework provides a robust means to create motion scenes. The users can move objects through a scene with whatever means they typically use. Then, by recording standard ROS published information, the algorithm has enough data to recreate these scenes in simulation. The user can also explore different scenes using the same recorded information without the need to modify the recorded data.

### 4.2 Modular Formats

Another goal for the tool is to allow arbitrary label formats. If a user desires a new labeling scheme to support some new network, they should be able to quickly write the specific code they need to parse the information, without concern for the rest of the system. This increases the usefulness of the tool across a range of use cases.

To accomplish this, the package uses a plugin style scheme. Specifically, the package defines a base class for all potential format writers using the Abstract Base Classes (ABC) library [19]. ABC enables the declaration of abstract methods that any inheriting class must implement. Each new format writer is implemented as an inheriting class, thus ensuring that each format writer interfaces correctly with the main algorithm. There are three specific methods that each inheriting class must implement. The main algorithm keeps a list of all the classes and calls the appropriate methods for each one at the correct time.

The first called method occurs at each successive step through the scene, when the objects are at their recorded poses for that instance in time. The complete scene information, including object poses, the raw image, and pixels masks for each object are all passed to the method. The method is then only responsible for extracting the specific information the format needs and writing it to file in the correct manner.

The next is a function called at the very end of the entire execution. This allows the format writers to perform any final steps required by the format. For instance, Darknet uses a main text file that contains a list of each image in the dataset. The final abstract method is one simply used to indicate if a particular format requires instance segmentation. As discussed later, this is only to help improve processing time.

The base class also offers two helper functions that are of use to most formats. It offers a method that transforms points between frames of reference and another that projects points into an image. A common helper function avoids the need to reimplement this transform functionality for each writer.

By defining the format writers in this manner, users can add new formats quickly and without reprogramming the core data collection process. For example, adding a Darknet format involves only a few steps. First, is the scene writing method. When the method is called, the Darknet format writer uses the provided information about the objects, including their location and vertices of bounding shapes, to transform the vertices into the image. From there, the format writer simply determines the pixel bounding box using these projected vertices and writes this bounding box to the right file. At the end of the execution, the Darknet format writer then writes the main list. Because the base class and main algorithm manages most functionality, this Darknet module is around 100 lines of code with no step more complicated than calling functions or finding an average. Implementing this functionality took considerably less time than implementing it in Version 1.

Following this approach, there are currently three formats defined within the package. They are Darknet [14], a network implementation known as PVNet [20] that uses projected keypoints, and a custom format that uses the vertices projected into an arbitrary frame of reference.

By using this modular format structure, users can easily add additional formats without unnecessary rewriting of the core functionality of data collection. This allows users to extend this tool to new formats without the need for extensive rework.

### 4.3 Instance Segmentation

An entire subclass of networks performs instance segmentation on image data. This requires labeled data that features unique identifiers for each instance of an object class and indicates which pixels contain an object in a given image. To support labeling data for these networks, this algorithm generates pixel masks to indicate which pixels belong to each object. The general algorithm is as shown below. **Fig. 1**. Shows an example observation generated by the algorithm, along with the raw image of the scene.

```
Instance Segmentation Initialization:
subtractor <- OpenCV.CreateBackgroundSubtractorMOG2()
Remove all objects from scene
For a user specified number of times:
  image <- capture image
  subtractor.apply(image)
Return subtractor
```

```
Instance Segmentation:
If instance segmentation required:
  For each object under control:
    store pose
    move out of camera view
  For each object under control:
    move back to pose
    image <- capture image
    mask <- subtractor.apply(image)
    mask <- apply image post-processing(mask)
    store mask for use by data writers
    Move out of camera view
```

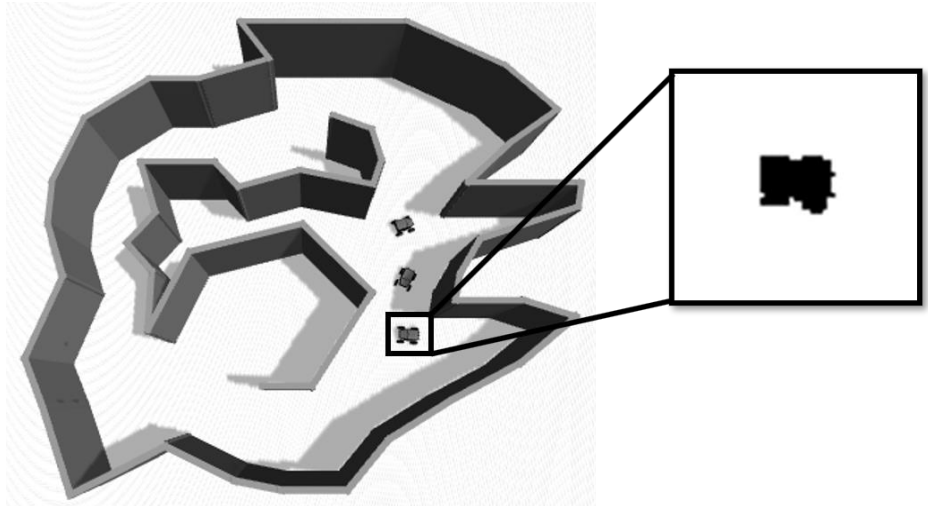

**Fig. 1.** A sample raw image collected during the data generation process and associated pixel mask for one of the objects in the scene. The colors of the mask are inverted for readability.

Prior to any steps in the above algorithm, the tool polls each format writer. If no formats require instance segmentation, the tool skips this entire routine. This helps increase processing speed by avoiding unnecessary image manipulation.

During initialization, the algorithm first creates a background subtractor. It uses OpenCV's Gaussian Mixture-based implementation [17]. A few of the parameters of the subtractor can be user specified to tune performance. It then moves every object out of view of the camera to capture background images. The algorithm collects multiple images to account for noise. As each image is captured, the tool applies them to the background subtractor to generate a reference background.

The initialized subtractor is then used during scene generation. At each step, the tool moves all objects out of view of the camera. Then, one at a time, it moves them back to their correct position within the scene. It then captures the image and runs it through the background subtractor. Because the object is the only thing in the scene besides the background, the results are a pixel mask indicating which pixels belong to the object. This process continues for each object. After capturing each object's pixel mas, the algorithm then moves all objects back into the scene in their correct spots to capture the raw image for that scene.

Right after image collection, the algorithm performs some post processing on the pixel masks. Using the known 3D bounding shape of the object and the correct transforms into the image, it constructs a bounding box around each object. It then uses this box as a filter for the pixel mask. No pixel outside of the box is set in the mask, since the box defines a conservative outer bound on possible pixels associated with the object. This helps reduce noise in the pixel mask.

When the tool calls each format writer to label a single scene, it provides these generated pixel masks to the writers. Currently, layering the masks to correctly match model occlusion in the scene remains an open question and is left to each data writer to manage. While this is an important functionality to include in the core algorithm or data writer base class, the authors chose to defer implementation due to project constraints as none of the formats currently in use by the team require a layered pixel map. However, one potential approach is to use the object's Euclidian distance to the camera to decide the ordering layer for the combined mask.

With this capability, the automatic generation of synthetic data can now extend to even more types of neural networks.

### 4.4 Dependencies

The last goal with the package was to limit the number of dependencies. During development, the team carefully selected which packages to use based on what the average ROS data scientist is likely to have installed. **Table 1** shows the dependencies. Almost all utilized packages come with the standard ROS installation and the default Python installation. The only two that do not, OpenCV [17] and NumPy [21], are used for background subtraction, image handling, and array manipulation. While these are not default packages, the team felt that any researcher working with image data is likely to already have these packages installed. By limiting these dependencies, the tool is easier to integrate into anyone's workflow.

**Table 1.** A list of dependencies required for this package. Bold dependencies do not come with the default Python and ROS installation.

| ROS Packages | Python Packages |
|---|---|
| cv_bridge | abc |
| gazebo_msgs | **cv2** |
| geometry_msgs | **numpy** |
| rosbag | os |
| rospy | |
| sensor_msgs | |
| tf2_ros | |

## 5    Usability

The previously discussed goals for this tool ensure that it is straightforward to use for anyone familiar with ROS. In general, a user will need to configure the initial setup, run the application, wait for it to finish, then start using the results. Each step only involves a few actions, if any, to complete. Additional steps are needed if new formats need incorporated.

### 5.1    Setup

While setting up the package for a specific job is the most involved step, it is still straightforward. The package comes with a detailed README that walks the user through the steps. Additionally, example files document all settings and describe their default behavior. Customizing these settings and completing setup involves three main steps.

First, the user must generate the necessary bag file. They can do this at any time prior to running the package. As discussed above, there are very few constraints for this bag file. The user need only record the TF tree while they move objects in the desired motion patterns. This is a standard procedure for any ROS user, so most users will already be familiar with this step.

After creating the bag file, the user must also create a simulated world in Gazebo. While in most cases, this world will mimic the one used to generate the bag file, that is not necessary. This environment can be as simple as an empty world or as complicated as a realistic office setup. The user must place a camera for image capture somewhere within the environment. They must also instantiate a number of objects representative of the objects controlled during bag file creation. The object identifiers must match the ones used when creating the bag file. However, visual appearances can differ.

Once the user records the bag file and has the Gazebo world is running, the last setup step is configuring the parameters. ROS uses YAML files to configure parameters from a single file. This is a standard way of customizing ROS packages. The package supplies a default YAML file to use as a template. It also has documentation on each parameter, many of which have default values if not provided. These parameters specify things such as which bag file to use, the list of objects to control, and where to place output files. Each data writing format may have its own parameters as well that the user can specify in the same YAML file.

After configuration of the parameters, the data generation is ready to run.

### 5.2 Execution and Performance

After completing setup, the user is ready to run the program. This is a simple, one line command to start the entire data generation process. After starting, the user need only wait while the program executes.

Starting the program uses a ROS concept called launch files, which is a standard way to start programs with user specified parameters. When started, the program checks the various user specified values. It ensures that each required parameter is set and warns the user if an optional parameter is unspecified. During execution, the package will periodically print the percent complete to the command line to update the user. No user interaction is required during execution.

During development, the authors measured performance of the generation tool. They explored runtimes across different numbers of objects, collections of data writers, and time. **Fig. 2** features a summary of results. There were two types of formats considered, one that requires instance segmentation and one that does not. The authors measured these results using Ubuntu's time command on a laptop with an i9 processor and 32 GB of memory.

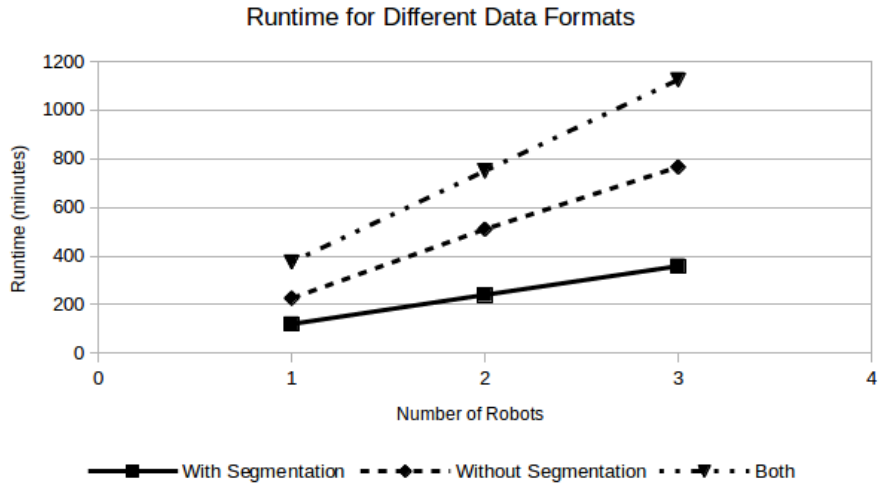

**Fig. 2.** A comparison of runtimes across different numbers of objects and different format writers.

As currently designed, this package is not fast. To promote modularity, some efficiency was sacrificed. For example, depending on the format specifications, multiple format writers might have to manipulate the data in the same way. The result is a running time that scales with the number of data writers. Additionally, the tool moves each object at each iteration. This increases the running time as the number of objects increase, as shown in **Fig. 2**.

The figure also shows the relative impact of performing instance segmentation. The format that requires it grows at a faster rate as the number of robots increase. This is due to the background subtractor. As written, it runs once for each robot in a scene. However, the main algorithm also queries each format writer and skips this step if not needed, resulting in the faster execution rate seen in the figure.

While the data generation time is lengthy, it is important to note that it is fully automated. Once the algorithm begins to collect data, no user intervention is necessary. This is one of the primary benefits to this package.

### 5.3    After Generation

Once execution is complete, the user has one or several complete sets of labeled data. This data is already in the correct format for training and can be used straight away without further user effort required. The only limited post processing a user might need is splitting out a test or validation set.

## 5.4 New Formats

When a new data format is required, the user must develop the code to write the format correctly. Following the above-mentioned structure, the user must implement a new class with a few required methods. Then, the user includes this new format class in the list of formats to call during execution. The included README documents this entire process.

To start making a new format, the user defines a new Python class. This class should inherit the base class offered by the package. Because this base class has abstract methods, the user must implement them and will receive an error if they do not and try to run the code anyway.

Within this new class, the user should initialize any parameters that the format requires. The base class already looks up the user specified output location, but the inheriting class can expand this to include any configuration settings needed. Next, the user implements the method called at each scene. This is the primary method used to write data to file. Each time it is called, the method is provided a list of all objects, their locations, object specific information such as keypoints, and image data. This method should use the provided information to translate positional information into appropriate labels. It should also write the necessary label files, as dictated by the format. After that, the user implements a final method called at the end of data generation. This provides an opportunity for any cleanup operations or writing metainformation, such as lists of label file locations. Lastly, the user should specify if the format requires instance segmentation pixel masks. As discussed previously, this allows the main algorithm to reduce computational time if no format requires segmentation.

After defining the class, the user then includes it in the library and list of classes called by the algorithm. These are both single line steps. Inclusion in the library involves importing the class into the overall data format module. Adding it to the tool involves instantiating an object within the main program. The documentation illustrates these steps using previous examples. After completion, the main code will automatically call the correct methods at the appropriate time.

By creating the data writer class, defining the methods, and including it in the main tool, the user can quickly add new data formats. Creating these new definitions does not require changing the main code. By including this functionality, the usefulness of the tool can continue to expand as the library of existing formats grows.

## 5.5 Vignette

As an example of the efficiency of the tool, consider the following anecdote. The authors were exploring a new network. They used this tool to generate data for a simple setup just to ensure the network was working. After creating the data, they found that the camera placement and resolution did not have sufficient coverage of specific parts of the scene. The authors adjusted the camera information in simulation and still used the same settings and recorded bag file to generate new data. The entire process of creating a new dataset took approximately 10 minutes of human effort compared to an estimated 1-2 hours for manual labeling. The authors then repeated this process a few times with new setups and simulated frame rates with similar levels of effort required. Additionally, the authors explored a new type of network. The authors quickly wrote up an additional plugin to write data for this new format. They completed the plugin and integrated it into the existing list with the others within an hour. Data generation then proceeded as usual. The result was a smooth data generation process that allowed the authors to spend minimal effort to create the data and more time exploring network optimization.

## 6  Remaining Questions

This synthetic data generation tool is a flexible approach to generating data for a range of neural network formats. However, there are still future improvements that would further enhance useability. This includes increasing speed, expanding the format library, and GUI development.

As discussed above, the algorithm suffers from poor scalability. While modularity typically introduces some overhead, code changes can likely improve the runtime. For example, some of the underlying algorithms can be rewritten to store and manipulate data more efficiently. As many routines are called multiple times, any performance gain is likely to have significant impact on overall runtime.

Additionally, expanding the data format library increases useability. Since the tool allows easy development of format writers, users can quickly incorporate new formats into the available list. This has the added benefit of testing underlying assumptions on the algorithm to ensure it is truly useable across a wide range of networks. The authors are already exploring Mask-RNN [22] and format writers to mimic common dataset formats, such as the Common Objects in Context (COCO) dataset [23].

Lastly, the inclusion of a GUI simplifies setup. While the current approach follows familiar ROS conventions, a well-designed GUI offers a means to guide the user through the setup process and address potential issues prior to running the tool. This would reduce errors and further decrease setup time.

The primary benefits of this tool are its simplicity and minimal user effort. Increasing the runtime, adding additional formats, and developing a GUI all contribute to these benefits and make the tool more useful.

## 7  Conclusion

This paper proposes an automated tool for generating annotated image training data for various object detection networks. The proposed package is the first to use ROS and Gazebo for the purpose of automatically generating synthetic annotated machine learning training data. Flexibility in package parameters like the simulated world, recorded motion plans, and camera parameters allows for rapid generation of diverse datasets with minimal effort from users. Additionally, the ability to easily add more data format plugins allows users to tailor the package to their specific data generation needs. Using this toolset, the authors created three fully annotated datasets in three separate formats with about 15 minutes of manual setup and several hours of hands-off running time, depending on the specific configuration. This package drastically reduces hands-on data collection and labeling time while ensuring the generation of accurate data in a modular way to allow various use cases.